\pdfoutput=1

\documentclass[11pt]{article}

\usepackage{ACL2023}

\usepackage{times}
\usepackage{latexsym}

\usepackage[T1]{fontenc}

\usepackage[utf8]{inputenc}

\usepackage{microtype}

\usepackage{inconsolata}

\usepackage{amsmath,amsfonts,bm}

\def\eqref#1{equation~\ref{#1}}

\def\1{\bm{1}}

\def\vb{{\bm{b}}}

\def\vs{{\bm{s}}}

\def\vx{{\bm{x}}}

\def\mO{{\bm{O}}}

\def\mS{{\bm{S}}}

\def\mW{{\bm{W}}}

\DeclareMathAlphabet{\mathsfit}{\encodingdefault}{\sfdefault}{m}{sl}
\SetMathAlphabet{\mathsfit}{bold}{\encodingdefault}{\sfdefault}{bx}{n}

\newcommand{\R}{\mathbb{R}}

\usepackage{hyperref}
\usepackage{url}
\usepackage{graphicx}

\usepackage{microtype}
\usepackage{xcolor}
\usepackage{fancyhdr}

\usepackage{times}
\usepackage{latexsym}

\usepackage{CJKutf8}
\usepackage{booktabs}
\usepackage{graphicx}
\usepackage{subfigure}
\usepackage{comment}
\usepackage{xspace}
\usepackage{adjustbox}
\usepackage{mathtools}
\usepackage{bbm}
\usepackage{multirow}
\usepackage{amssymb}
\usepackage{amsmath}
\usepackage{verbatim}
\usepackage{amsfonts}
\usepackage{bm}
\usepackage[normalem]{ulem}
\usepackage{wrapfig}
\usepackage{bbding}

\title{Emergent Modularity in Pre-trained Transformers}

\author{Zhengyan Zhang$^{1*}$, Zhiyuan Zeng$^{1*}$, Yankai Lin$^{2,3}$, Chaojun Xiao$^{1}$, Xiaozhi Wang$^{1}$\\ 
\textbf{Xu Han$^{1}$, Zhiyuan Liu$^{1,4,5\dagger}$, 
Ruobing Xie$^{6}$, Maosong Sun$^{1,4\dagger}$, Jie Zhou$^{6}$}\\
\textsuperscript{\rm 1}NLP Group, DCST, IAI, BNRIST, Tsinghua University, Beijing \\
\textsuperscript{\rm 2}Gaoling School of Artificial Intelligence, Renmin University of China, Beijing \\
\textsuperscript{\rm 3} Beijing Key Laboratory of Big Data Management and Analysis Methods \\       
\textsuperscript{\rm 4}International Innovation Center of Tsinghua University, Shanghai\\
\textsuperscript{\rm 5}Quan Cheng Laboratory \textsuperscript{\rm 6} Pattern Recognition Center, WeChat AI, Tencent Inc \\ 
\texttt{\{zy-z19,zengzy20\}@mails.tsinghua.edu.cn} \texttt{\{liuzy,sms\}@tsinghua.edu.cn}\\
}

\begin{document}
\maketitle
\begin{abstract}
This work examines the presence of modularity in pre-trained Transformers, a feature commonly found in human brains and thought to be vital for general intelligence.
In analogy to human brains, we consider two main characteristics of modularity: 
(1)~\textit{functional specialization of neurons}: we evaluate whether each neuron is mainly specialized in a certain function, and find that the answer is yes.
(2)~\textit{function-based neuron grouping}: we explore finding a structure that groups neurons into modules by function, and each module works for its corresponding function.
Given the enormous amount of possible structures, we focus on Mixture-of-Experts as a promising candidate, which partitions neurons into experts and usually activates different experts for different inputs. 
Experimental results show that there are functional experts, where clustered are the neurons specialized in a certain function.
Moreover, perturbing the activations of functional experts significantly affects the corresponding function.
Finally, we study how modularity emerges during pre-training, and find that the modular structure is stabilized at the early stage, which is faster than neuron stabilization. It suggests that Transformers first construct the modular structure and then learn fine-grained neuron functions. 
Our code and data are available at \url{https://github.com/THUNLP/modularity-analysis}.

\end{abstract}

\section{Introduction}

{\let\thefootnote\relax\footnotetext{$^*$ Equal contribution}}
{\let\thefootnote\relax\footnotetext{$^\dagger$ Corresponding authors}}

Recently, pre-trained Transformers have shown the potential to achieve general intelligence~\citep{GPT-3,fei2022towards,gato,GPT-4,SparkAGI}, which encourages researchers to explore the analogy between Transformers and human brains~\citep{DBLP:conf/nips/TonevaW19,DBLP:conf/icml/CaucheteuxGK21,caucheteux2022brains,goldstein2022brain}.
These works have shown that the behaviors of Transformers resemble those of human brains. Do the internal structures of Transformers also mirror those of human brains in order to achieve similar behaviors?

Neuroscientists have found that the structure of neuron organization in human brains follows a modular pattern~\citep{bullmore2009complex,meunier2010modular}, which has two main characteristics.
(1)~\textbf{Functional specialization of neurons}: each neuron is mainly specialized in a certain function.
(2)~\textbf{Function-based neuron grouping}: neurons with the same function are clustered in a local region, and each function relies on a specific region.
In this work, we wonder whether Transformers also organize neurons in a modular way.

(Q1) \textbf{Are neurons functionally specialized?} 
The functional specialization of neurons is the basis of modularity.
To this end, we propose a novel framework to analyze the functionality of neurons in Transformers.
In this framework, we study three representative functions by a unified method, including semantic function~\citep{scarlini-etal-2019-just,DBLP:journals/corr/abs-2005-07647}, knowledge function~\citep{DBLP:journals/tacl/JiangXAN20,DBLP:conf/acl/DaiDHSCW22}, and task function~\cite{SkillNeuron}.
Experimental results show that the neurons in pre-trained Transformers become much more specialized than those in randomly-initialized ones after self-supervised learning on large-scale corpora.
Moreover, in pre-trained Transformers, there are several groups of neurons, each of which excels in a specific function.

(Q2) \textbf{Is there a modular structure of neurons?} 
In analogy to human brains, a modular structure should group neurons with the same function together as modules, and each module plays a crucial role in a specific function.
Since there are thousands of neurons in a Transformer, it is impractical to iterate over all possible structures of neurons.
We consider the grouping of Mixture-of-Experts~(MoE)~\cite{Jacobs1991-zn,Switch-Transformer} as a promising candidate, which gracefully partitions neurons into experts and is widely used in Transformers.
Moreover, most MoE models are sparsely activated, which is similar to human brains.
Specifically, we study two types of MoE models, pre-partitioned MoE (pre-MoE) and post-partitioned MoE (post-MoE). 
Pre-MoE refers to the model architectures that expand feedforward layers by MoE to improve model capacity before pre-training~\citep{Switch-Transformer}.
Post-MoE refers to the models that are converted from vanilla Transformers to their MoE version by partitioning feedforward layers into experts after pre-training~\citep{zhang2022moefication}. 
It provides a way to group neurons without changes in parameters and forward pass.

By studying the function distribution on experts, we find that both pre-MoE and post-MoE Transformers have a strong tendency to distribute the neurons excelling in a certain function concentratedly into some experts.
Moreover, perturbing the activations of 3\% of experts by function leads to performance close to random guessing, which is more significant than perturbing the same amount of individual neurons specialized in the function. 
Therefore, the MoE structure indeed reflects the modularity of pre-trained Transformers.

(Q3) \textbf{How does modularity emerge during pre-training?} 
By analyzing the pre-training process, we find that the functions of expert networks are stabilized to a large extent at the early stage (around 15\% of the total training steps) for both pre-MoE and post-MoE Transformers, which is faster than the neuron stabilization (around 75\% of the total steps).
Our findings provide evidence for a coarse-to-fine mechanism of pre-training, which first constructs the coarse modular structure and then learns fine-grained neuron functions.

\section{Related Work}

\textbf{Interpreting Pre-trained Transformers.} 
Although pre-trained Transformers have achieved great success in the field of NLP~\citep{DBLP:journals/corr/abs-2111-01243,DBLP:journals/corr/abs-2108-07258}, their inner working mechanism is still a black box.
Researchers explore interpreting the pre-trained Transformers~\citep{DBLP:journals/tacl/RogersKR20} from different perspectives, such as hidden representations~\citep{DBLP:conf/naacl/Liu0BPS19}, attention matrices~\citep{voita-etal-2019-analyzing,DBLP:conf/blackboxnlp/ClarkKLM19}, and output distributions~\citep{DBLP:conf/emnlp/PetroniRRLBWM19}.
Among them, studying the behavior of a single neuron is an important branch~\citep{DBLP:journals/corr/RadfordJS17,DBLP:journals/corr/abs-2108-13138,bills2023language}, which is most related to our work.
From the neurons in feedforward layers, researchers have found various encoded information such as concepts, facts, and task abilities~\citep{DBLP:journals/corr/abs-2005-07647,DBLP:conf/acl/DaiDHSCW22,SkillNeuron}.
In this work, we study how these neurons are organized to form a modular structure, which is a new interpretation perspective.

\textbf{Modularity of Neural Networks.}
Modularity is a widespread property in complex systems, both artificial~\citep{DBLP:conf/aaai/Ballard87,baldwin2000design} and biological~\citep{von1999modularity,lorenz2011emergence,clune2013evolutionary}.
Previous work mainly focuses on incorporating explicitly designed modules into neural networks~\citep{DBLP:conf/cvpr/AndreasRDK16,DBLP:conf/nips/KirschKB18,mittal2020learning,DBLP:conf/iclr/GoyalLHSLBS21,mittal2022modular,mittal2022compositionalattention}.
Recently, \citet{DBLP:journals/corr/abs-2110-08058,DBLP:conf/iclr/CsordasSS21,dobs2022brain} study whether standard neural networks naturally manifest modularity without deliberate design, and have discovered some naturally-emerging modular structures of CNNs and LSTMs.
Compared to previous work, we extend the modularity analysis to pre-trained Transformers, whose capabilities are expected to be stronger.

\textbf{Transformers with MoE.}
Sparse MoE is usually used to enlarge the model capacity of Transformers while keeping the computational efficiency~\citep{GShard,baselayers,Switch-Transformer}.
Specifically, for a given input, MoE conditionally selects a subset of experts to process the input and then combines the outputs of these experts to generate the final output.
Beyond computation efficiency, MoE is also used to implement modular Transformers~\citep{DEMix,SkillNet,DBLP:conf/naacl/PfeifferGLLC0A22,DBLP:journals/corr/abs-2208-10442}.
These works explicitly design extra constraints during pre-training to ensure the modularization of expert networks.
However, it is still unclear whether modular experts can emerge naturally in Transformers during pre-training.

\section{Functionality Evaluation}
\label{sec:evaluation}
In this section, we first introduce the definition of neurons and how to evaluate the functionality of neurons and experts.
Then, we briefly introduce the evaluation setups, including the pre-trained models.  

\textbf{Neurons in Transformer.}
Following~\citet{DBLP:conf/acl/DaiDHSCW22,SkillNeuron}, we study the neurons in the feedforward networks (FFNs), which account for about two-thirds of Transformer parameters.
Previous work has shown that there is rich information in FFNs~\citep{DBLP:journals/corr/abs-2005-07647,geva-etal-2021-transformer,geva2022transformer,DBLP:conf/acl/DaiDHSCW22,SkillNeuron}.

Specifically, the FFN is a two-layer MLP and computes the output by $\text{FFN}(\vx) = \mW^O\sigma(\mW^I\vx + \vb^I) + \vb^O$, where $\mW^I \in \R^{d_{ff}\times d}, \mW^O \in \R^{d\times d_{ff}}$ are the weight matrices, $\vb^I \in \R^{d_{ff}}, \vb^O \in \R^{d}$ are the bias vectors, $d$ is the dimensions of input and output, $d_{ff}$ is the dimensions of intermediate hidden states, and $\sigma$ is the activation function. 
For simplicity, we discard the bias terms in the following part.
For fine-grained analysis, we dissect the FFN into neurons and rewrite the FFN equation as 
\begin{equation}
\label{eq:ffn}
\small
\text{FFN}(\vx) = \sum_{i=1}^{d_{ff}} \sigma(\mW^I_{i,:} \cdot \vx) \mW^O_{:,i},
\end{equation}
where $\mW^I_{i,:}$ and $\mW^O_{:,i}$ are the $i$-th row and column of $\mW^I$ and $\mW^O$, respectively.
The FFN output is the sum of the outputs of all neurons.
From this perspective, we define a neuron $n_i$ as a row vector $\mW^I_{i,:}$ and a column vector $\mW^O_{:,i}$.
The neuron activation of $n_i$ is $\sigma(\mW^I_{i,:} \cdot \vx)$.
The number of neurons in an FFN is equal to the intermediate hidden dimension $d_{ff}$.

Sparse Mixture-of-experts in Transformer is a variant of FFN~\citep{GShard}, which significantly increases the model capacity by adding more parameters and keeps the computational cost affordable.
In MoE layers, each expert is an FFN, and the output of the MoE layer is the weighted sum of the outputs of all experts.
We can also define neurons of MoE as Equation~\ref{eq:ffn}.
The neuron-based form of MoE is provided in Appendix~\ref{sec:moe-neuron}.

\textbf{Predictivity for Functions.}
To comprehensively study the functions of neurons and experts, we cover three typical functions. %
For each function, we construct various sub-functions, each of which is a fine-grained version of the function. %
There are 576 sub-functions in total. Here we introduce the three functions and their sub-functions.

\textit{Semantic Function.} 
Semantic function refers to the ability to understand the meaning of input texts.
In this work, we focus on how neurons capture the patterns of word senses.
We use a large-scale dataset with word-sense annotations, OneSec~\citep{DBLP:conf/acl/ScarliniPN19}, to construct binary classification data for semantic sub-functions.
In OneSec, each sentence has a keyword whose sense is annotated based on Wikipedia\footnote{\url{https://en.wikipedia.org/}}.
We first filter out the keywords that have more than one sense in the dataset and then randomly select 100 sentences for each sense.
For each sense pair of a word, we construct a binary classification dataset by labeling the sentences with one sense as positive and the sentences with the other sense as negative.
Finally, we have 529 semantic sub-functions, each of which is a binary classification problem to distinguish the two senses of a word.
For experiments on Switch Transformer in Section~\ref{sec:pre-training}, we randomly sample 100 semantic sub-functions for evaluation.

\textit{Knowledge Function.}
Knowledge function refers to the ability to memorize factual knowledge.
In this work, we focus on the factual triples, which are used to construct knowledge graphs.
We define a knowledge sub-function as a binary classification to identify whether a triple is correct.
Specifically, we sample several triples from Wikidata as positive instances and randomly replace their head or tail entities to construct negative instances.
We group these instances according to their relations and each relation has its corresponding knowledge sub-function.
There are 39 knowledge sub-functions from T-REx~\cite{trex,DBLP:journals/tacl/ElazarKRRHSG21} and each sub-function has 400 instances.

\textit{Task Function.}
Task function refers to the ability to perform downstream tasks.
Previous work has shown that training a small part of parameters in pre-trained Transformers can achieve comparable performance to full-parameter fine-tuning~\citep{DBLP:conf/emnlp/LesterAC21,DBLP:conf/iclr/HuSWALWWC22} so that Transformers are supposed to learn amounts of task knowledge from pre-training.
In this work, we use several classification datasets. from GLUE~\citep{wang2018glue}, including SST-2~\citep{socher2013recursive}, QQP\footnote{\url{https://data.quora.com/First-Quora-Dataset-Release-Question-Pairs}}, MNLI~\citep{williams2018broad}, CoLA~\citep{warstadt2018neural}, MRPC~\citep{dolan2005automatically}, RTE~\citep{dagan2006pascal}, QNLI~\citep{rajpurkar2016squad}.
There are 8 task sub-functions in total because MNLI is split into two binary classification tasks.
To stimulate these sub-functions, we adopt the input templates provided by~\cite{T5} to improve neuron predictivity.
For each label class, we randomly sample 1000 instances for evaluation if the number of instances under this class is larger than 1000.

Admittedly, coarse is our function classification. It does not cover all functions learned by pre-trained Transformers, and there are interactions between each pair of functions so there is unavoidable overlap.
However, we focus on a unified framework and concrete evaluation approach, and they can be easily generalized to other ways of function classification, meaning that our contribution is independent of function classification. Using this way of classification is simply due to its typicality.

To evaluate the ability of a neuron to capture the pattern of a sub-function, we compute the predictivity of the neuron activations for the sub-function.
Following~\citet{DBLP:journals/corr/abs-2005-07647,SkillNeuron}, we focus on the sub-functions that can be formulated as a binary classification problem.

\begin{figure*}
\centering
\includegraphics[width=\linewidth]{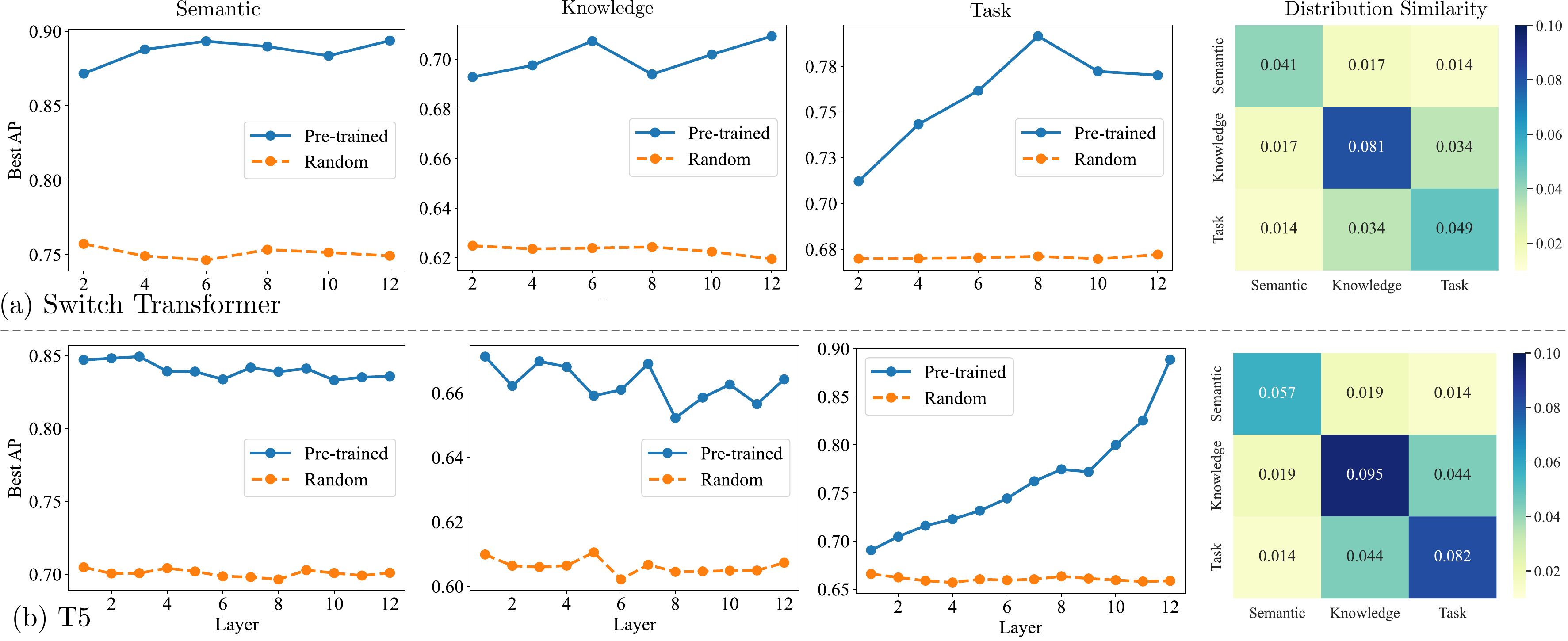}
\caption{Left part: average predictivity of each function of each layer for the pre-trained models and their randomly-initialized counterparts of (a) Switch Transformer and (b) T5. Right part: average distribution similarity between different functions of different layers for the same pre-trained models. All of the similarities of random-initialized models are around $0.01$, including the diagonal elements, which are significantly lower than those of the pre-trained models.
}
\label{fig:func-sim}
\end{figure*}

We denote the dataset of a sub-function as $\mathcal{D} = \{(s_i, y_i)\}_{i=1}^{\lvert\mathcal{D}\rvert}$, where $s_i$ is the input sequence and $y_i \in \{0, 1\}$ is the label.
Since the computation of FFNs is word-wise, we define the activation of the neuron $n_j$ of a sequence $s_i$ as $a_{ij} = \max_{\vx_k \in \vs_i} \sigma(\mW^I_{j,:} \cdot \vx_k)$, where $\vs_i = \{\vx_1, \vx_2, \ldots, \vx_l\}$ is the hidden states of $s_i$ and $l$ is the length of $s_i$.
Then, we have the pairs of neuron activations and labels, $\mathcal{A}_j = \{(a_{ij}, y_i)\}_{i=1}^{\lvert\mathcal{D}\rvert}$.
Based on $\mathcal{A}_j$, we compute the average precision (AP)~\cite{zhu2004recall} of the neuron activations as the predictivity of the neuron $n_j$.
For each expert, we compute the average AP of all neurons in the expert as the predictivity of the expert for the sub-function.
Please refer to Appendix~\ref{sec:exp-details} for more details.

\textbf{Evaluation Setups.}
We evaluate two kinds of MoE models, pre-partitioned MoE and post-partitioned MoE.
(1)~\textbf{Pre-partitioned MoE} expands feedforward layers by MoE before pre-training.
Switch Transformer~\citep{Switch-Transformer} is a representative pre-MoE model.
The architecture of Switch Transformer is similar to T5~\citep{T5} except that Switch Transformer replaces FFNs in the even Transformer layer with MoE layers.
(2)~\textbf{Post-partitioned MoE} converts a pre-trained vanilla Transformer into its MoE model by partitioning FFNs into experts.
Since \citet{zhang2022moefication} show that vanilla Transformers have implicit MoE structures by discovering the inner correlation among neurons, we use the same method to MoEfy vanilla Transformers.
Note that we only adopt MoEfication to provide a neuron grouping. We do not change any architecture, parameters, forward function (computing all neurons and then summing them up), etc, making the MoE-fied T5 \textbf{identical} to the original T5.

Due to the computational cost, we consider the Switch Transformer with 16 experts and use the same number of experts for the post-MoE models.
Since the functionality evaluation does not involve decoding, we only compute the neuron and expert predictivities of the encoders.
Besides, we focus on the neurons in MoE layers for Switch Transformer to facilitate the modular analysis in the following sections.

\section{Functional Specialization of Neurons}
\label{sec:neuron}
In this section, we analyze the functional specialization of neurons.
First, we compare the neuron predictivities of pre-trained models with those of randomly-initialized models at the layer level, which will present a general picture of neuron functions.
Second, we study how functions distribute on neurons in each layer.

\textbf{Neuron predictivities of different layers.} At each layer, we compute the best predictivity of neurons for each sub-function and then calculate the average best predictivity among all sub-functions in each function.
For comparisons, we also evaluate randomly-initialized models.
We report the results in the left part of Figure~\ref{fig:func-sim}. 

From this figure, we have two observations.
(1)~The average best predictivities of pre-trained neurons are significantly higher than those of randomly-initialized neurons, indicating that the neurons have learned these functions from pre-training and the neurons with top-ranked predictivities indeed excel in corresponding sub-functions.
(2)~The best predictivity of the task function increases with the layer number while the best predictivities of the semantic and knowledge functions vary little across layers.
It suggests that the task function may be more difficult than the semantic and knowledge functions so the higher layers are more suitable for learning the task function.
Note that we evaluate the frozen pre-trained models on the tasks without fine-tuning or prompt-tuning and the pre-training data of Switch Transformer do not contain the tasks. Hence, we can exclude the possibility of optimization artifacts~\cite{DBLP:conf/acl/DurraniSD21}, which may make the higher layers more related to the task function.

\textbf{Distribution in each layer.}
We first identify the neurons with the top predictivity rankings for each sub-function as \textit{sub-functional neurons} in each layer and then compute the overlap between the two sets of sub-functional neurons.
Formally, assuming that we identify the top $k$ neurons for each sub-function, the overlap score is defined as $\frac{\sum_{n_i \in \mathcal{N}_1} \mathbb{I}(n_i \in \mathcal{N}_2)}{k}$,
where $\mathcal{N}_1$ and $\mathcal{N}_2$ are the sets of neurons for the two considered sub-functions.
If the overlap score is high, there is a group of neurons that are good at both sub-functions.
In the experiments, we consider the neurons with the top 1\% predictivities for each sub-function.
Since there are hundreds of sub-functions, it is impossible to display all of them in a figure and we compute the average overlap score between two functions to measure the distribution similarity between different functions.
Note that we omit the self-overlap scores, which are always equal to 1.
The results are reported in the right part of Figure~\ref{fig:func-sim}.

\begin{table*}[]
\begin{center}
\small
\begin{tabular}{lclllllll}
\toprule
\multirow{2}{*}{Model}                      & \multirow{2}{*}{Pre-trained} & \multirow{2}{*}{Partitioning}       & \multicolumn{2}{c}{Semantics} & \multicolumn{2}{c}{Knowledge} & \multicolumn{2}{c}{Task}  \\
& & & Prop. & Degree & Prop. & Degree & Prop. & Degree\\
\midrule
\multirow{3}{*}{Switch Transformer} & \XSolidBrush & Random & 0.002 & 0.037 & 0.001 & 0.024 & 0.001 & 0.018 \\
& \Checkmark & Random  &   0.226     &  1.038    & 0.052 & 0.652& 0.003& 0.066\\ 
& \Checkmark                      & Pre-MoE          & \textbf{0.354}      &  \textbf{1.490}    &  \textbf{0.260}& \textbf{1.560}& \textbf{0.219}& \textbf{1.604}\\ 
\midrule
\multirow{3}{*}{T5} & \XSolidBrush & Random & 0.016 & 0.257 & 0.001 & 0.031 & 0.001 & 0.017\\        
& \Checkmark  & Random      &  0.252      &  1.203    & 0.061 & 1.031& 0.007 & 0.221\\ 
& \Checkmark& Post-MoE &    \textbf{0.338}     &  \textbf{2.000}    & \textbf{0.214} & \textbf{2.686} & \textbf{0.120} & \textbf{3.276}\\ \bottomrule
\end{tabular}
\end{center}
\caption{Proportion of functional experts to all experts and their modularization degree.
Higher modularization degrees mean that functional experts have more sub-functional neurons than the expectation of uniform distribution.
}
\label{tab:stat-test}
\end{table*}

From this figure, we observe that:
(1)~In the pre-trained models, the distribution similarity of the same function is significantly larger than that of different functions, which indicates that there are some groups of neurons, each of which is good at a certain function.
And, it is not mainly caused by the similarity that naturally exists between sub-functions of the same function because the results of randomly-initialized models do not show such a phenomenon as shown in Appendix~\ref{sec:random}.
Hence, we conclude that \textbf{there are some emergent groups of functionally-specialized neurons after pre-training.}
(2)~One neuron may be capable of multiple sub-functions even from different functions. 
For example, the average overlap score between the knowledge function and task function is also significantly higher than that of random models so there are some neurons good at both knowledge and task sub-functions.

\section{Finding Modular Structure of Neurons}
\label{sec:expert}
Since MoE is a promising candidate for the modular structure of Transformers, we analyze the MoE structure in this section.
First, we verify whether the neurons specialized in a certain function are concentrated in some experts, called functional experts.
Second, we perturb the activations of experts corresponding to a certain function. Third, we train the model on different datasets, retaining only functional experts by removing non-functional experts. The last two experiments verify the importance of the experts for their corresponding functions.

\textbf{Distribution on experts.} If experts were not functionally specialized, the sub-functional neurons would be randomly distributed among experts.
Hence, we conduct statistical hypothesis testing to evaluate whether the neuron distribution on experts is significantly different from random.
Assume that there are $N$ neurons in a layer, $n_E$ neurons in each expert, $k$ sub-functional neurons for each sub-function, and $M$ sub-functions in a certain function.
The null hypothesis is that the sub-functional neurons of each sub-function are independently and randomly distributed on experts, i.e., the number of sub-functional neurons in each expert follows a hypergeometric distribution with parameters $N$, $K$, and $n_E$.
The sum of the numbers of sub-functional neurons on all sub-functions in an expert is denoted by $r_i$.\footnote{We do not find a general form for the distribution of the sum of independent hypergeometric distributions. Since $K$ is significantly smaller than $N$, we approximate the hypergeometric distribution with a binomial distribution.}
The alternative hypothesis is that an expert has a larger $r_i$ than expected by chance.

In the experiments, we also treat the neurons with the highest $1\%$ predictivities for each sub-function as its sub-functional neurons.
For each function, we compute the p-value of the sum of the hypergeometric distribution for each expert and reject the null hypothesis if the p-value is less than $0.001$.
We also conduct the same experiment on random partitioning, where the neurons are randomly partitioned into expert-sized clusters, and randomly-initialized counterparts.
We regard the experts that reject the null hypothesis as \textit{functional experts} and report the proportion of functional experts to all experts in each function.
And, we also consider the modularization degree, which is defined as the relative ratio of functional neurons in the expert compared to a uniform distribution.
The proportion of functional experts and the modularization degree are computed by
\begin{equation}
\small
\text{Prop.}=\frac{E_f}{E}, \quad \text{Degree} = \frac{r_i}{n_E}/(\frac{Mk}{N}),
\end{equation}
where $E_f$ is the number of functional experts, $E$ is the number of experts, $\frac{r_i}{n_E}$ is the proportion of functional neurons in the expert, and $\frac{Mk}{N}$ is the proportion expectation of functional neurons under a uniform distribution.
The overall degree is $0$ if no functional expert exists, and otherwise is the average degree among all functional experts. 

The results are shown in Table~\ref{tab:stat-test}.
From this table, we have three observations. 
(1)~Since the sub-function distributions of pre-trained models are not independent, the proportion of functional experts of pre-trained models with random partitioning is higher than that of random-initialized counterparts.
(2)~However, the proportion of functional experts of pre-trained models with MoE partitioning is still higher than that with random partitioning.
Moreover, the modularization degree of the functional experts in MoE structures is significantly higher than that in random partitioning.
It indicates that the experts of both pre-MoE and post-MoE are more likely to intensively include neurons excelling in a certain function.
(3)~We further compare the predictivities of the functional experts and non-functional experts and find that the functional experts have significantly higher predictivities than the non-functional experts in their corresponding functions. It indicates that our quantification for expert predictivities is consistent with the concept of functional experts.
More details are in Appendix~\ref{sec:Predictivity_of_functional_experts}.

\begin{figure}
\centering
\includegraphics[width=\linewidth]{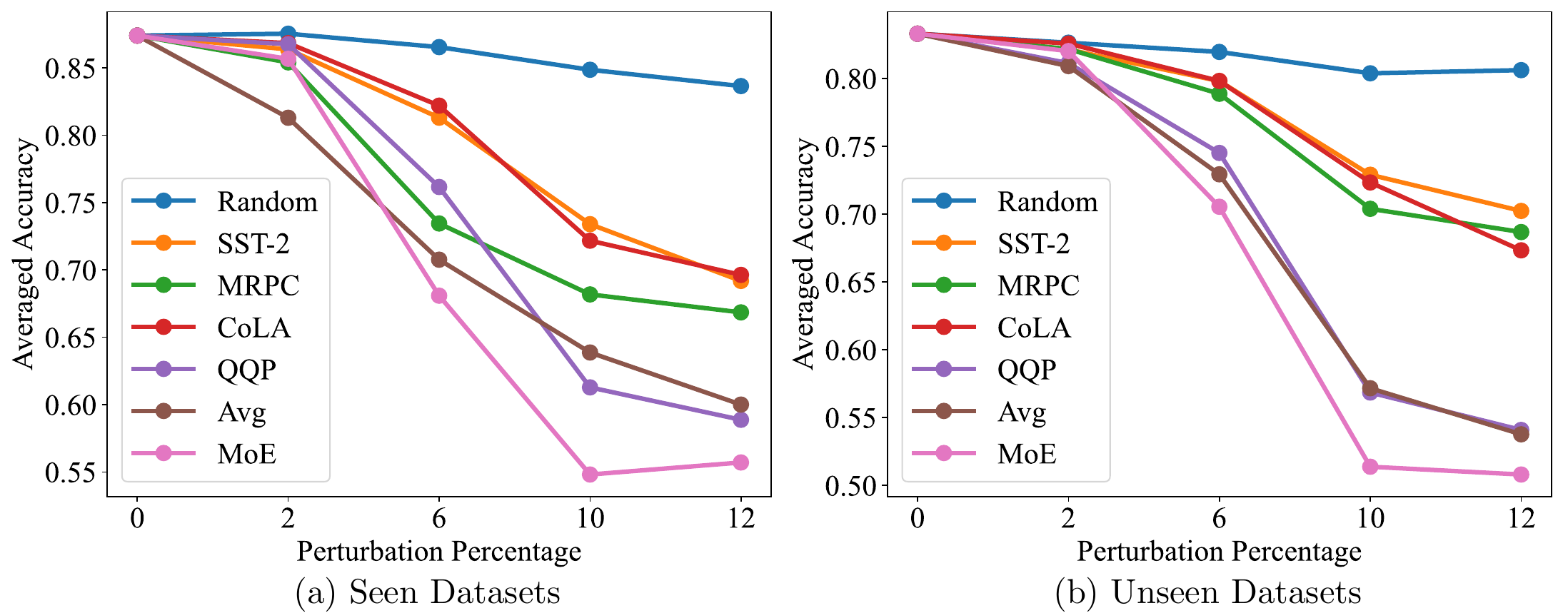}
\caption{Perturbation performance of T5. For ``Random'', we randomly perturb neurons. For ``SST-2'', ``MRPC'', ``CoLA'', ``QQP'', we are guided by the neuron predictivities on each dataset and perturb the top-ranked neurons. For ``Avg'', we sum the predictivities on all datasets above and also perturb the top-ranked neurons. For ``MoE'', we consider the experts with top-ranked sums of the predictivities. The seen datasets are the four datasets above. The unseen datasets include five other datasets.}
\label{fig:perturbation}
\vspace{-1em}
\end{figure}

\textbf{Perturbation analysis.} Furthermore, we conduct perturbation experiments, which are widely used to analyze both biological and artificial neural networks~\citep{Cowley2022.07.18.500505,SkillNeuron}, to evaluate the causal effect of experts on model performance.
Since T5 is a dense model and Switch Transformer is a sparse model, we use different perturbation methods for them. Specifically, the pre-MoE models only select one expert at each layer, so we choose to perturb the selection function. For the post-MoE models, which select all experts at each layer, we perturb the activation values of the targeting experts. Due to the difference in experiment setups, the perturbation performance is also different and their perturbation results are not comparable.

For T5, we perturb all neuron activations of the target experts by adding random noises to them and evaluate the perturbed models on the downstream tasks.
We rank experts according to their sum of the predictivities for several downstream datasets and perturb the top-ranked experts.
We regard the datasets used in computing the sum of the predictivities as seen datasets, including SST-2~\cite{socher2013recursive}, MRPC~\cite{dolan2005automatically}, CoLA~\cite{warstadt2018neural}, and QQP.
To evaluate the generalization ability of the experts, we also perturb them and evaluate the perturbed models on unseen datasets, including MNLI~\citep{williams2018broad}, QNLI~\citep{rajpurkar2016squad}, CB~\citep{demarneffe:cb}, MultiRC~\citep{khashabi2018looking}, and BoolQ~\citep{clark2019boolq}.
We compute the average performance of the perturbed models on the seen and unseen datasets, respectively.
For comparisons, we also conduct neuron-level perturbation and keep the proportion of perturbed neurons equal to that of expert-level perturbation.
There are three kinds of neuron-level perturbations: (1)~perturb the neurons that have top-ranked predictivities for a certain dataset, (2)~perturb the neurons that have top-ranked sums of the predictivities for seen datasets, and (3)~perturb the neurons randomly.
We perturb the neurons in the last four layers where the task function is mainly located as shown in Figure~\ref{fig:func-sim}.
For fine-grained perturbation percentage, we partition FFNs into $96$ experts, which is different from that of the other experiments.

\begin{figure*}
\centering
\includegraphics[width=\linewidth]{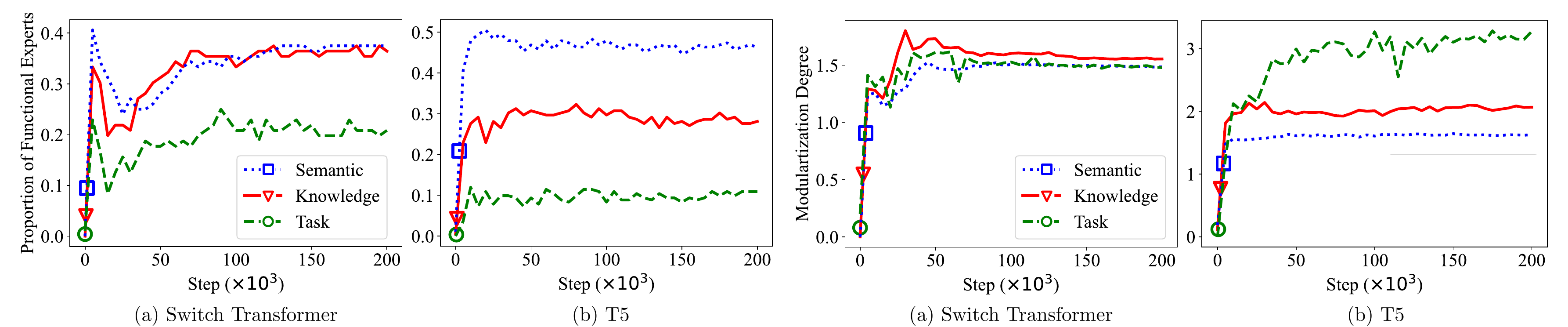}
\caption{Changing curves of the proportion of functional experts and their modularization degree. We also mark the value of the last checkpoint with random partitioning on the curve by points with different shapes.}
\label{fig:emergence}
\vspace{-1em}
\end{figure*}

We report the average accuracy of the perturbed models on the downstream tasks in Figure~\ref{fig:perturbation}.
From this figure, we have three observations.
(1)~The experts with high predictivities are very important for the model performance.
For example, perturbing $10\%$ of the experts by function in the last four layers (around $3\%$ of the total experts in the model) decreases the average accuracy by nearly $30\%$ and makes the model perform as random guessing.
(2)~For neuron-level perturbations, ``Avg'' perturbation achieves a larger performance drop than single-dataset perturbation, which is expected because intuitively it perturbs the neurons with high overall predictivities.
(3)~\textbf{Perturbing experts leads to a more significant performance drop than perturbing individual neurons} on both seen and unseen datasets when the perturbation proportion is higher than $6\%$.
It suggests neurons in the experts cooperate instead of working independently so perturbing them will influence the cooperation and lead to a significant performance drop.
We conclude single neurons can not perform a function well in lack of the cooperation with modules despite their high overall predictivity.

\begin{table}
\centering
\small
\begin{tabular}{lrrr}
\toprule
& Original & No Function & Function \\
\midrule
7-task Avg. & 73.23 & 72.78 & \textbf{74.45} \\
\bottomrule
\end{tabular}
\caption{Perturbation performance of Switch Transformer on GLUE tasks.}
\label{tab:switch}
\vspace{-1em}
\end{table}

For Switch Transformer, in the last four layers, we perturb the selection probabilities of experts.
We constraint the model to select from a subset of experts by setting the selection probabilities of the other experts to $0$ and fine-tune the perturbed models on the downstream tasks with the template in~\citet{LM-BFF}.
There are two kinds of perturbations: 
(1)~No Function: we only select the non-functional experts in a certain task, which is equivalent to setting the activations of the functional experts to $0$.
(2)~Function: we only select the functional experts in a certain task.
Since the number of functional experts is smaller than that of non-functional experts, we randomly select a subset of non-functional experts for No Function to make the size of the subset equal to that of the functional experts.
Besides, we also report the performance of the original model without any perturbation.
More details are in Appendix~\ref{sec:exp-details}.
From Table~\ref{tab:switch}, we observe that avoiding the functional experts leads to an overall performance drop.
Furthermore, only selecting the functional experts even achieves a higher performance than the original model.

In summary, we observe that the specialized neurons tend to be located concentratedly in some experts based on function and the functional experts play an important role when the model performs related functions.
Hence, it is reasonable to study the modularity of Transformers by MoE.

\section{Emergence of Modularity}
\label{sec:pre-training}
In this section, we study the emergence of modularity during pre-training.
To this end, we pre-train the base version of T5 and Switch Transformer from scratch.
We use the MoEfication partitioning of the last checkpoint as the MoE structure of T5.
More details of pre-training are in Appendix~\ref{sec:exp-details}.

\textbf{Emergence Patterns of Functional Experts.} We first study the changing curves of the proportion of functional experts and their modularization degree during pre-training, i.e., we apply the same analysis in Section~\ref{sec:expert} to each checkpoint.
The results are shown in Figure~\ref{fig:emergence}.
We have the following observations. (1)~The proportion of functional experts and their modularization degree quickly achieves a high point, and then keeps relatively stable till the end.
It indicates that functional experts emerge at the early stage of pre-training.
(2)~The proportion of functional experts in the Switch Transformer fluctuates significantly at about 20K steps and its stabilization is slower than that of T5.
It suggests that the emergence of the modular structure in Switch Transformer is surprisingly more difficult than T5.
The reason may be that Switch Transform omits the gradients of unselected experts, which causes the optimization to be harder than that of T5~\citep{DBLP:conf/icml/DuHDTLXKZYFZFBZ22,DBLP:journals/corr/abs-2202-08906}. %

\begin{figure*}
\centering
\includegraphics[width=0.9\linewidth]{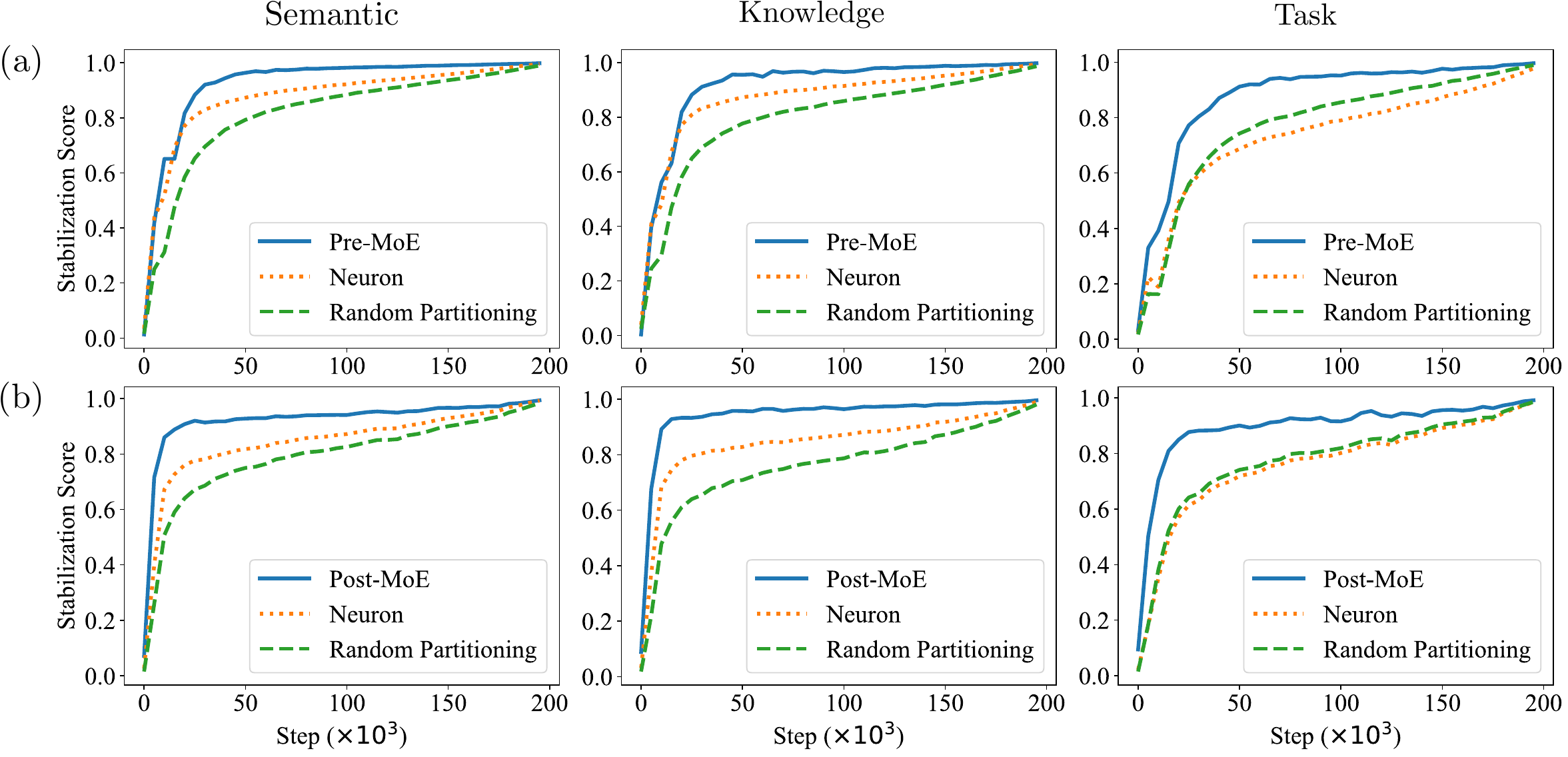}
\caption{Spearman's rank correlation between the functionality distributions of two adjacent checkpoints.}
\label{fig:stabilization}
\vspace{-1em}
\end{figure*}

\textbf{Stabilization of Experts and Neurons.}
Even though clear is the changing curve of the number and modularization degree of functional experts from a global perspective, we still do not know how the predictivities of neurons and experts changes.

There are two kinds of predictivity dynamics. 
The first is the changing of the absolute predictivities, and the second is the relative order changing of predictivities among all experts or neurons.
While it is straightforward to study the absolute predictivities, it is difficult to have a consistent analysis standard due to different scales for different functions and layers. 
Hence, we focus on the relative order changing of predictivities.
Intuitively, some experts or neurons may excel in a sub-function at a certain stage, and maintain this relative dominance as pre-training continues.
With this in mind, we study the stabilization of predictivity rankings.

To study the stabilization of predictivity rankings, we quantify the similarity between a layer of two model checkpoints w.r.t. a particular sub-function, which is either at the expert level or at the neuron level. Specifically, for a sub-function, we define such a similarity as Spearman's rank correlations~\citep{Spearman} between the predictivities of experts or neurons in the considered layer of the two checkpoints. In this way, we measure to what extent the predictivities of the two checkpoints is positively correlated.
We measure the similarity between two adjacent (saved) checkpoints as stabilization score, which reflects the trend toward stabilization. Higher similarity indicates a lower changing pace and thus a higher degree of stabilization. For each function, we show the curve of average stabilization score among all sub-functions in it and across all layers, both at the expert level and neuron level. 
To facilitate our analysis, we also measure it on random partitioning.

We report the result in Figure~\ref{fig:stabilization}. From this figure, we have four observations. (1)~During the pre-training, both experts and neurons are increasingly stabilized. (2)~Experts are stabilized to a large extent at the early stage of pre-training. It takes around 15\% of the total training steps for the expert predictivities to achieve a stabilization score of 0.9. (3)~Expert stabilization is notably faster than both neuron stabilization (around 75\% of the total training steps) and the stabilization for random partitioning\footnote{While the numbers of experts and neurons are different, the correlation scores are comparable~\cite{de2016comparing}.}. In conclusion, we see strong evidence that \textbf{coarse-to-fine is the inner mechanism of pre-training}.
Transformer first learns a modular structure, where the structure becomes stable at the early stage, and then there is a fine-grained process to learn neuron functions.

\section{Discussion}

\textbf{Efficient Pre-training.}
Large-scale pre-training of Transformers is very expensive~\citep{GPT-3}.
It is promising to use MoE to reduce the computational cost by activating only a small part of the experts.
Our findings have shown the emergent modularity of experts, which demonstrates the reasonableness of the MoE structure.
However, as shown in Section~\ref{sec:pre-training}, Switch Transformer is more unstable than T5 on the modular structure at the beginning.
It suggests that we should begin with a dense model and gradually make it sparse instead of directly training a sparse model from scratch, which has been explored in some preliminary works~\citep{DBLP:journals/corr/abs-2112-14397,DBLP:conf/nips/HazimehZCSCMHC21}.

\textbf{Model Fusion.}
Considering there are many pre-trained models on different corpora and even modalities, researchers have started to explore how to fuse them to aggregate different knowledge together.
Compared with model ensembling, model fusion has the potential to be more efficient because it does not compute all of the models.
Much of the current research in model fusion focuses on weight averaging and achieves some promising results~\citep{DBLP:journals/corr/abs-2208-03306,DBLP:journals/corr/abs-2111-09832}.
However, weight averaging requires two models having the same architecture, which is not always the case.
In this work, we discover the modular structure of pre-trained Transformers, which may facilitate the model fusion based on module combinations, which gets rid of the architecture constraint.

\textbf{Connection between Brains and Pre-trained Transformers.}
Building an artificial brain that corresponds to the human brain is an important neuroscience problem, e.g., the Blue Brain project~\citep{BlueBrain}.
Currently pre-trained Transformers show strong power for predicting brain signals~\citep{DBLP:conf/nips/TonevaW19,DBLP:conf/icml/CaucheteuxGK21}, but more fine-grained connections between the two are still not clear.
In analogy to brain regions, we present the modular structure of pre-trained Transformers.
It will be interesting to explore the connection between brain regions and the Transformer modules in the future.

\section{Conclusion}
In this paper, we study the modularity of pre-trained Transformers and validate the emergence of modularity.
We also study the pre-training process to understand the emergence of modularity and find the coarse-to-fine mechanism of pre-training.
We expect our evaluation framework and findings will facilitate and inspire future research in this area.

\section*{Limitations}
The major limitations of our work are as follows:
(1)~We show that the neuron structure of MoE reveals the presence of modularity in pre-trained Transformers.
However, the MoE structure is not the only possible modular structure.
To better understand the modular structure of Transformers, we need to explore more types of structures.
For example, the number of neurons in each module could be different, and the modular structure could be hierarchical, where modules are grouped into larger modules.
(2)~We study three typical functions for language processing: semantic function, knowledge function, and task function.
There are many other functions that could be studied, such as the syntactic function, discourse function, etc.
Moreover, our categorization of functions may be not suitable for pre-trained Transformers because there are some overlaps between studied functions.
A new Transformer-based function categorization may be needed.
(3)~We transform T5 into its MoE version to study its modular structure while not all dense pre-trained Transformers can be studied in this way because the adopted MoEfication technique~\citep{zhang2022moefication} can only transform ReLU-based Transformers.
Studying the modularity of other dense pre-trained Transformers, such as BERT, is also important for future research.

\section*{Acknowledgements}

This work is supported by the National Key R\&D Program of China (No. 2020AAA0106502) and Institute Guo Qiang at Tsinghua University.

\paragraph{Author Contributions} Zhengyan Zhang and Zhiyuan Zeng wrote the code and conducted the experiments. Zhengyan Zhang and Zhiyuan Zeng wrote the initial draft. Yankai Lin, Chaojun Xiao, Xiaozhi Wang, Xu Han, and Zhiyuan Liu, Ruobing Xie significantly edited and improved the paper. Maosong Sun and Jie Zhou provided valuable advice to the research.

\bibliography{anthology}
\bibliographystyle{acl_natbib}

\appendix
\newpage
\section{Appendix}

\subsection{Neurons of MoE}
\label{sec:moe-neuron}

In MoE layers, each expert is an FFN, and the output of the MoE layer is the weighted sum of the outputs of all experts, $\text{MoE}(\vx) = \sum_{i=1}^E \alpha_i \text{FFN}_i(\vx)$, where $\alpha_i$ is the weight of the $i$-th expert and $\text{FFN}_i$ is the $i$-th expert.
$\alpha_i$ is computed by a gating network.
Note that the weights of unselected experts are zero, which makes the MoE layer sparse.
We can also rewrite the MoE layer into a neuron-based form, 
\begin{equation}
\small
\begin{aligned}
\text{MoE}(\vx) &= \sum_{i=1}^E \alpha_i \sum_j \sigma(\mW^I_{i,j,:} \cdot \vx) \mW^O_{i,:,j} \\
    &= \sum_{i,j} \sigma(\mW^I_{i,j,:} \cdot \vx) \alpha_i \mW^O_{i,:,j},
\end{aligned}
\end{equation}
where $\mW^I_{i,j,:}$ and $\mW^O_{i,:,j}$ are the $j$-th row and column of $\mW^I_i$ and $\mW^O_i$ in the $i$-th expert, respectively.
The gating weight $\alpha_i$ is non-negative and can be viewed as the scaling factor of $\mW^O_{i,:,j}$.
Correspondingly, we define a neuron $n_{i,j}$ as a row vector $\mW^I_{i,j,:}$ and a column vector $\mW^O_{i,:,j}$, and the neuron activation of $n_{i,j}$ is $\sigma(\mW^I_{i,j,:} \cdot \vx)$.

\begin{figure*}
    \centering
    \includegraphics[width=\linewidth]{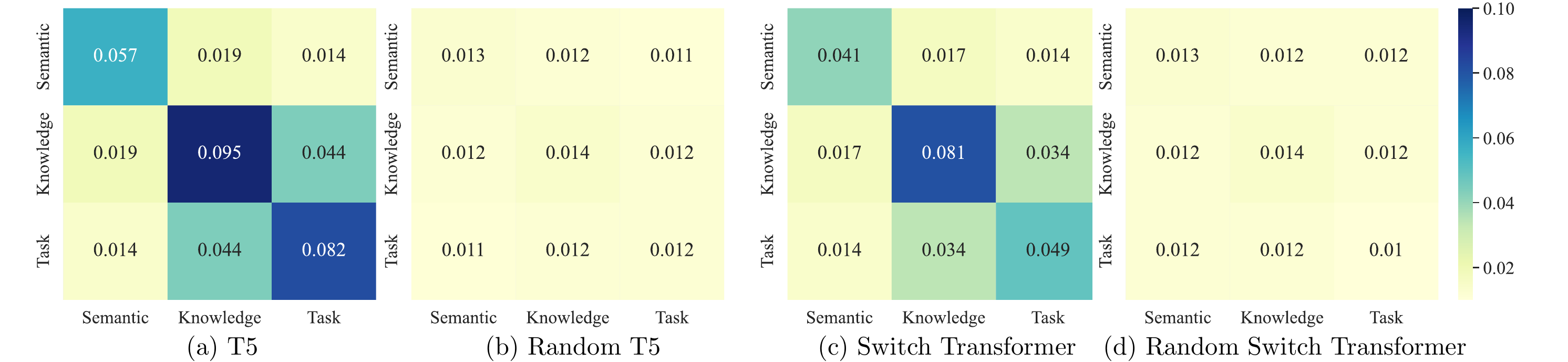}
    \caption{Distribution similarity between different sub-functions. We report the average similarity between functions. We consider the pre-trained models and their randomly-initialized counterparts.}
    \label{fig:func-sim-2}
\end{figure*}

\subsection{Experimental Details}
\label{sec:exp-details}

\textbf{Calculation of AP.} AP is the weighted average of precision at different recall levels, which is a common metric for evaluating the performance of binary classification models.
Since AP only represents the positive correlation, we compute the APs of both neuron activations and their opposite values, $-a_i$, and take the maximum as the final AP.
The final AP ranges from 0.5 to 1, where 0.5 means the neuron is useless for the sub-function and 1 means the neuron is perfect for the sub-function.

\textbf{Randomly-initialized models.} In Section~\ref{sec:neuron}, the evaluation of randomly-initialized models is conducted 3 times and we report the average results.

\textbf{Perturbation analysis on T5.} To match the magnitude of neuron activations of T5, we set the variance of the Gaussian noise to be 4.
The perturbation analysis is conducted 5 times and we report the average results.

\begin{table}
\centering
\tiny
\begin{tabular}{lrrr}
\toprule
& Original & No Function & Function \\
\midrule
CoLA & 72.11~($\pm$0.01) & 72.96~($\pm$0.01) & \textbf{73.14}~($\pm$0.01) \\
MNLI & 53.94~($\pm$0.03) & 50.94~($\pm$0.02) & \textbf{56.24}~($\pm$0.03) \\
MRPC & 75.83~($\pm$0.02) & 75.18~($\pm$0.02) & \textbf{77.24}~($\pm$0.02) \\
RTE & \textbf{65.66}~($\pm$0.04) & 65.30~($\pm$0.04) & 65.43~($\pm$0.03) \\
QNLI & 75.16~($\pm$0.02) & 76.19~($\pm$0.02) & \textbf{78.34}~($\pm$0.02) \\
QQP & 77.14~($\pm$0.01) & 76.80~($\pm$0.01) & \textbf{78.26}~($\pm$0.01) \\
SST-2 & \textbf{92.79}~($\pm$0.01) & 92.10~($\pm$0.00) & 92.53~($\pm$0.01) \\
7-task Avg. & 73.23 & 72.78 & \textbf{74.45} \\
\bottomrule
\end{tabular}
\caption{Perturbation performance of Switch Transformer on GLUE tasks. For each dataset, we report the mean and standard deviation (std) accuracy over eight random seeds, using the \texttt{mean} $\pm$ \texttt{std}.}
\label{tab:switch_full}
\vspace{-1em}
\end{table}

\begin{table*}[]
\begin{center}
\small
\begin{tabular}{llllllll}
\toprule
\multirow{2}{*}{Model}                      & \multirow{2}{*}{Partitioning}       & \multicolumn{2}{c}{Semantics} & \multicolumn{2}{c}{Knowledge} & \multicolumn{2}{c}{Task}  \\
& & Prop. & Degree & Prop. & Degree & Prop. & Degree\\
\midrule
Switch      & Random      &   0.001     &  0.022  & 0.001 & 0.022 & 0.001 & 0.022\\ 
Transformer & Pre-MoE     & \textbf{0.138}   &   \textbf{1.968}   & \textbf{0.101} & \textbf{2.028} & \textbf{0.124} & \textbf{2.029}\\ 
\midrule
\multirow{2}{*}{T5}        & Random       &  0.001      &  0.021 & 0.001 & 0.021 & 0.001 & 0.021\\ 
& Post-MoE &    \textbf{0.030}     &   \textbf{2.330}   & \textbf{0.038} & \textbf{2.985} & \textbf{0.039} &  \textbf{3.522} \\ \bottomrule
\end{tabular}
\end{center}
\caption{Proportion of sub-functional experts identified by hypothesis testing and their modularization degree. The result is averaged within each function.
}
\label{tab:stat-test-sub}
\end{table*}

\textbf{Perturbation analysis on Switch Transformer.} In this experiment, the predictivity and thus the functional experts are calculated based on the training set. Since Switch Transformer has not been trained on the considered datasets during pre-training, we train it on 128 randomly-picked training instances. We tune $\mW^O$ in the last four years as it takes the neuron activations as inputs, plus the routers in MoE layers. The learning rate and batch size are always 2E-4 and 16 respectively and we take Adam~\cite{DBLP:journals/corr/KingmaB14} as the optimizer. The epoch is 100, 300, 50, 100, 300, 150, and 200 for CoLA, MNLI, MRPC, RTE, QNLI, QQP, and SST-2. We run each experiment with 8 different seeds and report the average best performance on the validation set. The result for each dataset is reported in Table~\ref{tab:switch_full}. 

\textbf{Hyper-parameters of pre-training.} 
The pre-training corpus is OpenWebText~\citep{radford2019language}, which contains 40GB of web text.
We use the same pre-training task as the official T5 and Switch Transformer, which is masked language modeling.
The total number of training steps is 200K and we save the model every 5K steps.
We use the same hyper-parameters for pre-training to avoid the effect of hyper-parameters on our analysis, and it is also a common practice when comparing dense and sparse T5s~\citep{DBLP:journals/corr/abs-2202-08906}.
The learning rate is 1e-4. The batch size is 512. The max lengths of encoder inputs and decoder inputs are 512 and 256, respectively. We use 8 NVIDIA A100 GPUs for pre-training. The total pre-training time is around 3 days.

\textbf{Experiments on random partioning.} In Section~\ref{sec:expert} and Section~\ref{sec:pre-training}, we do the hypothesis testing on random partitioning, and we also calculate Spearman's rank correlation between adjacent checkpoints on random partitioning. These random experiments are done 1000 times and we report the average results.

\subsection{Function Distribution in Each Layer}
\label{sec:random}

Following Section~\ref{sec:neuron}, we report the distribution similarity of the randomly-initialized models in Figure~\ref{fig:func-sim-2}, which is significantly different from that of the pre-trained models.

\subsection{Predictivity of Functional Experts}
\label{sec:Predictivity_of_functional_experts}

We quantify the predictivity of the expert for sub-functions in Section~\ref{sec:evaluation} and define functional experts in Section~\ref{sec:expert}, and different are the technical details of quantification and definition. Hence, we conduct an experiment to check their consistency. For a function, we calculate $b_i$ as the average predictivity across all sub-functions for each expert $e_i$, and we also denote $f_i\in\{0,1\}$ as whether $e_i$ is a functional expert or not. Now we have $\mathcal{B} = \{(b_i, f_i)\}_{i=1}^E$, based on which we compute the AP. A high AP indicates a high consistency. We report the average AP across all layers in Table~\ref{tab:Predictivity_of_functional_experts}.

\begin{table}[]
\begin{center}
\small
\begin{tabular}{llll}
\toprule
\multirow{1}{*}{Model}                      & \multicolumn{1}{c}{Semantics} & \multicolumn{1}{c}{Knowledge} & \multicolumn{1}{c}{Task}  \\
\midrule
Switch Transformer  &   0.957     &  0.795    & 0.734 \\ 
\midrule
T5        &  0.912      &  0.894    & 0.931 \\ 
\bottomrule
\end{tabular}
\end{center}
\caption{Average AP calculated based on $\mathcal{B} = \{(b_i, f_i)\}_{i=1}^E$, where $b_i$ is the average predictivity across all sub-functions for expert $e_i$, and $f_i$ is whether $e_i$ is a functional expert or not.}
\label{tab:Predictivity_of_functional_experts}
\end{table}

The average AP is quite high. Therefore, we are confident that the quantification for expert predictivity is consistent with the concept of functional experts.

\subsection{Sub-Functioanl Experts}

\begin{figure*}
\centering
\includegraphics[width=\linewidth]{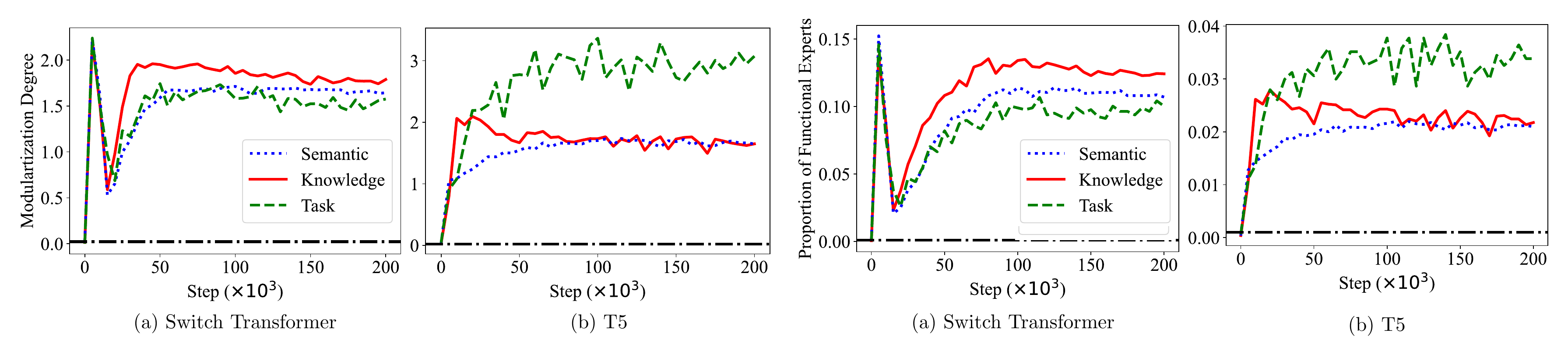}
\caption{Changing curves of the proportion of sub-functional experts and their modularization degree. The horizontal line is the value for random partitioning.}
\label{fig:sub-emergence}
\end{figure*}

Similar to the concept of functional experts discussed in Section~\ref{sec:neuron}, we can also define so-called sub-functional experts. Basically, for each sub-function, we conduct statistical hypothesis testing on its sub-functional neurons. We similarly calculate the proportion of sub-functional experts and their modularization degree. We report the average result within each function. The result of the Switch Transformer and T5 used in Section~\ref{sec:neuron} is reported in Table~\ref{tab:stat-test-sub}. The changing curve of the Switch Transformer and T5 trained by us is reported in Figure~\ref{fig:sub-emergence}.

\subsection{Organization of Sub-Functions}
We further study how the model organizes sub-functions into their functional experts\footnote{Strictly speaking, we did not define functional experts for a sub-function. In this context, the concept of ``functional experts'' is used to refer to the experts that have high predictivity for a sub-function.} and how the organization changes during the pre-training. From the perspective of sub-functions, it is basically how a sub-function shares functional experts with others.

For a function, we can list all the sub-functions within it denoted as $w_1,w_2,\dots,w_M$. The similarity score between each pair of sub-functions can be seen as a matrix $\mS$, where $\mS_{i,j}$ is the similarity score between $w_i$ and $w_j$ for all $1\leq i,j\leq M$. For a given $k$, we denote $\mO^{(k)}_{i,j}$ as the top $k$ expert overlap for $w_i$ and $w_j$. When Spearman's rank correlation between $\mS_{i,:}$ and $\mO^{(k)}_{i,:}$ (denoted as $V^{(k)}_i$) is high, it indicates that the sub-functions similar to $w_i$ share more functional experts than sub-functions dissimilar to $w_i$ do, and vice versa. According to the meaning, we call $V^{(k)}_i$ clustering score.

We do a case study on the semantic function of the Switch Transformer, which focuses on understanding word meanings.
Since relatively mature is the method of quantifying word similarity, we take $\mS$ as the word similarity matrix calculated by spaCy~\citep{spacy2}. Note that the features at the word level are the lowest level of semantic information, so the word similarity reflects the lowest level of similarity between two semantic sub-functions.

\begin{figure}
\centering
\includegraphics[width=0.8\linewidth]{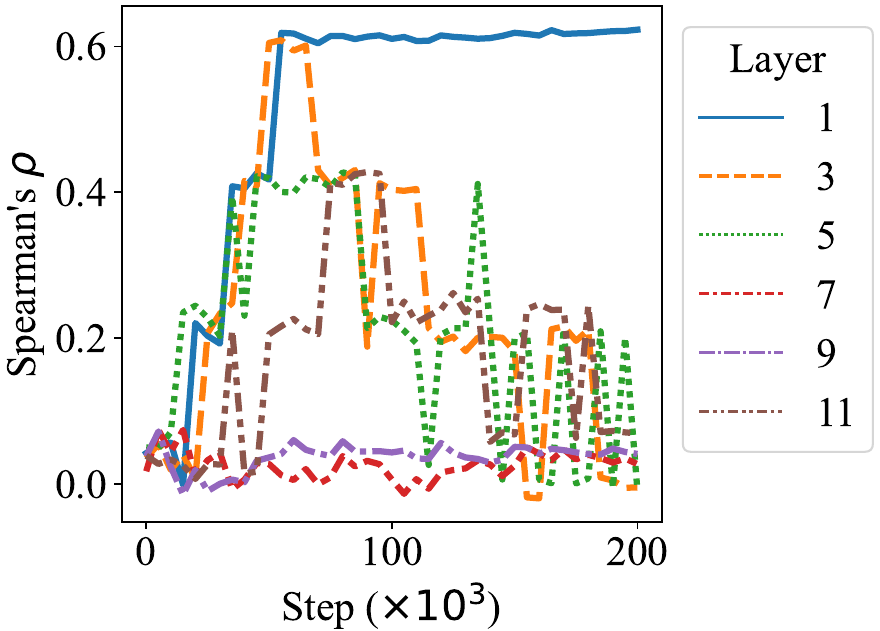}
\caption{Changing curve of the average clustering scores. We plot the curve for each MoE layer in Switch Transformer.} %
\label{fig:word_trend}
\vspace{-1em}
\end{figure}

\begin{figure}
\centering
\includegraphics[width=0.8\linewidth]{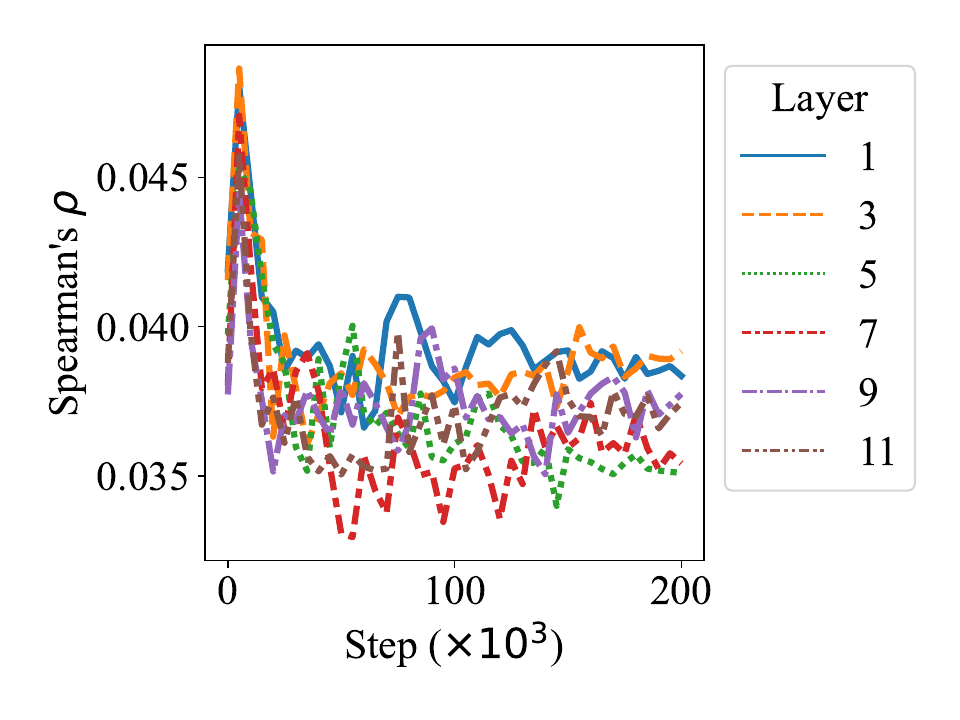}
\caption{Changing curves of the average clustering scores on random partitioning. We plot the curve for each MoE layer in Switch Transformer.}
\label{fig:word_trend_random}
\vspace{-1em}
\end{figure}

We report the curve of $\frac{\sum_{k=1}^K\sum_{i=1}^MV^{(k)}_i}{KM}$ for each layer in Figure~\ref{fig:word_trend}. We also report the result for random partitioning in Figure~\ref{fig:word_trend_random}. From this figure, we have the following observations. (1)~During pre-training, the clustering score of the lowest MoE layer (Layer 1) quickly achieves 0.6 and then keeps stable till the end. It proves that the pre-training tends to organize sub-functions sharing similar low-level information into the same functional experts in the low layer.
However, the final clustering score of higher MoE layers is close to 0, indicating that high layers do not organize sub-functions based on word similarity.
We guess that the reason is that the high layers may process high-level semantic information, which is not related to the word similarity.
(3)~We see three interesting curves of layers 3, 5, and 11. Their clustering scores achieve a high point when the clustering score of layer 1 first achieves its highest point, and then they continuously decrease to 0. The trend that high layers become increasingly responsible for high-level features may grow faster when the low-layer organization has been established than when the organization is forming.

\end{document}